\documentclass[letterpaper, 10 pt, conference]{ieeeconf}  

\IEEEoverridecommandlockouts                              

\usepackage{graphicx} 
\usepackage{amsmath} 
\usepackage{amsfonts,amssymb}
\usepackage[dvipsnames]{xcolor}
\usepackage{algorithm, algorithmic}
\usepackage{tabularx, booktabs, makecell, caption}
\usepackage{tabularray}
\usepackage{siunitx}
\usepackage{pifont}

\makeatletter
\def\algbackskip{\hskip-\ALG@thistlm}
\makeatother

\title{\LARGE \bf
Autonomous Behavior Planning For Humanoid Loco-manipulation Through Grounded Language Model
}

\author{Jin Wang$^{1}$$^{2}$$^{*}$, Arturo Laurenzi$^{1}$, Nikos Tsagarakis$^{1}$
\thanks{†This work was supported by the European Union’s Horizon 2020 research and innovation programme, euROBIN EPUE034001, and JL Leonardo ETCM058502.}
\thanks{$^{1}$Humanoids and Human-Centered Mechatronics (HHCM), Istituto Italiano di Tecnologia, Via Morego 30, Genoa, 16163, Italy.}
\thanks{$^{2}$DIBRIS, Universita di Genova, Italy, 16145.}
\thanks{$^{*}$Corresponding author: {\tt\small wang.jin@iit.it}}
}

\begin{document}

\maketitle
\thispagestyle{empty}
\pagestyle{empty}

\begin{abstract}
Enabling humanoid robots to perform autonomously loco-manipulation in unstructured environments is crucial and highly challenging for achieving embodied intelligence. This involves robots being able to plan their actions and behaviors in long-horizon tasks while using multi-modality to perceive deviations between task execution and high-level planning. Recently, large language models (LLMs) have demonstrated powerful planning and reasoning capabilities for comprehension and processing of semantic information through robot control tasks, as well as the usability of analytical judgment and decision-making for multi-modal inputs. To leverage the power of LLMs towards humanoid  loco-manipulation, we propose a novel language-model based framework that enables robots  to autonomously plan behaviors and low-level execution under given textual instructions, while observing and correcting failures that may occur during task execution. To systematically evaluate this framework in grounding LLMs, we created the robot ‘action’ and ‘sensing’ behavior library for task planning, and conducted mobile manipulation tasks and experiments in both simulated and real environments using the CENTAURO robot, and verified the effectiveness and application of this approach in robotic tasks with autonomous behavioral planning.
\end{abstract}

\section{INTRODUCTION}

Maintaining autonomy during the execution of a task in a real-world environment is both essential and challenging for robots, especially when performing tasks that require interaction with surroundings and manipulation of objects. This demands a high level of capability from robots that have to perceive and make decisions during the task execution and the ability to achieve autonomous planning based on these decisions. 
Furthermore, one of the main challenges lies in enabling robots to understand semantic instructions from humans and apply them within the context of different scenarios and their current state. This process involves encoding textual information into a hierarchical sequence of robot behaviors, as well as mapping high-level tasks to low-level robot control and generating reference trajectories that the robot can execute.

Integrating large language models (LLMs) has emerged as a promising avenue for enhancing the autonomy of robots. These models have demonstrated great flexibility in understanding and processing semantic information, along with remarkable reasoning and decision-making capabilities. However, due to the complexity of whole-body control and perceptual decision-making in humanoid robots, it is challenging to directly apply LLMs for action generation and planning, especially when it involves understanding the current task based on environmental cues and interacting with objects. Previous work has shown that incorporating language models into robotic tasks \cite{ahn2022can} and enabling intelligence to better reason and evaluate textual information can interact with the environment to accomplish long-horizon tasks that require complex planning of robot action sequences. Leveraging this feature, we design a language model-based planner, which requires the pre-creation of a $\textit{behavior lib}$ for the robot, including multiple actions and perceptions for different modalities. This can be used to generate direct reference trajectories for execution. After acquiring the human instruction and the semantic indices from the $\textit{behavior lib}$, the LLM generates hierarchical $\textit{task graphs}$, which guide the robot to follow the logical sequence of tasks and make decisions according to different scenarios.

\begin{figure}
\centerline{\includegraphics[width=9cm]{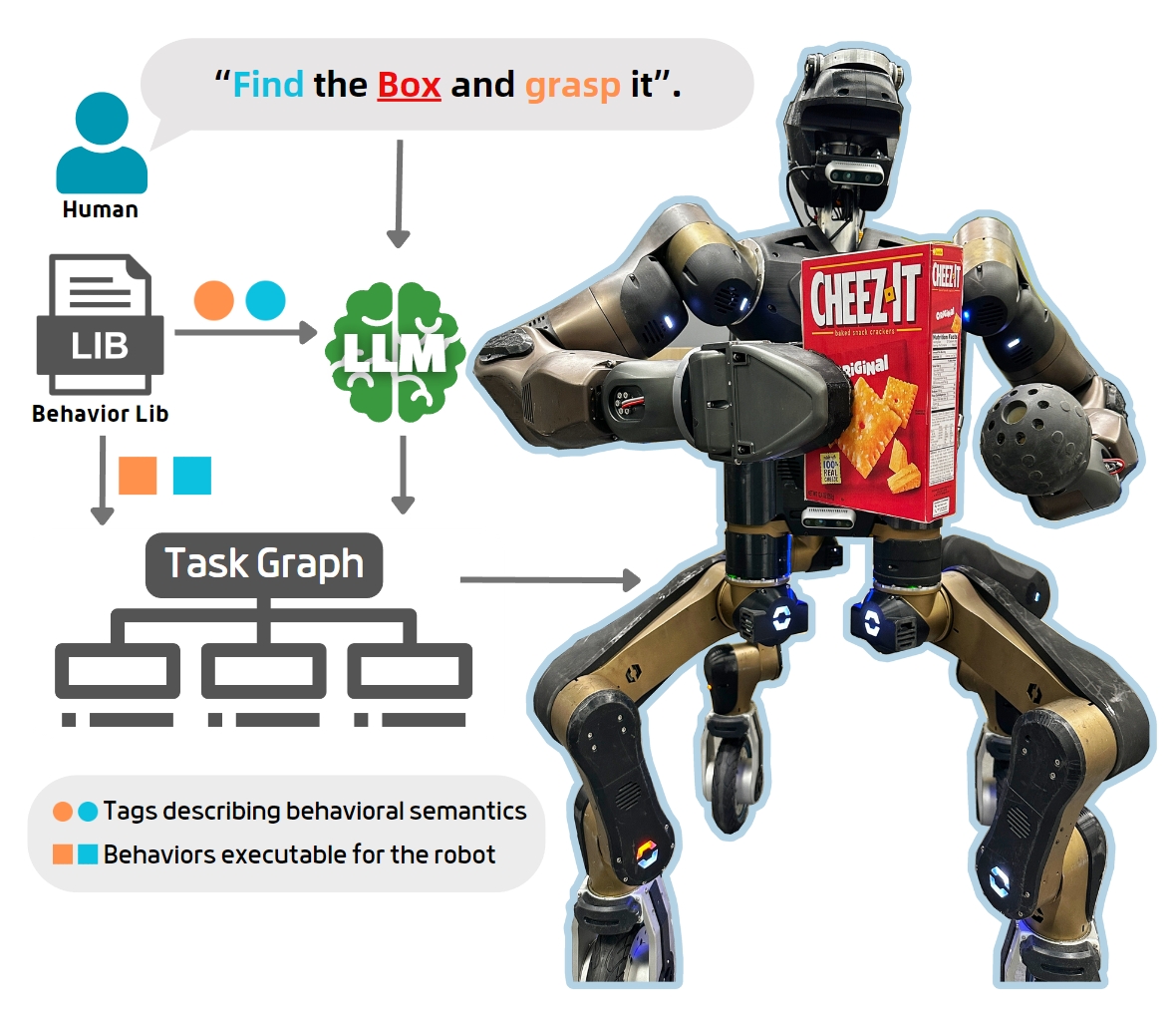}}
\caption{Humanoid robot CENTAURO picks objects with the planning of the $\textit{task graphs}$ generated by the LLM. The `$\textit{behavior lib}$' consists of various action and sensing behaviors with `tags' describing the semantic content of different behaviors. }
\label{fig1}
\end{figure}

Finally, the increasing autonomy of robots during the execution of complex tasks, as well as unexpected perturbations in long-term missions, make failures during the execution of high-level behavioral sequences in real-world environments inevitable. Therefore, failure detection as well as correction procedures, are needed to ensure alignment between low-level execution and high-level semantic planning during the programming of complex loco-manipulation tasks that are composed of multiple subtasks. We attempt to fuse multimodal sensor data and add them as perceptual behaviors into the robot $\textit{behavior lib}$, which enables the LLM to select the optimal combination of behaviors according to the task scenario. Among them, the visual language model (VLM) is used as one of the key metrics for determining whether the task is completed or not as well as for failure determination, due to its accurate and effective image information comprehension and reasoning capabilities. Once the VLM perceptual behavior detects a failure during a task, it triggers the behavioral planner to perform the corrective action. This process is pre-planned by the LLM and stored in the $\textit{task graphs}$, and by combining the perceptual behavior of different modalities, a closed loop of high-level feedback is realized, which improves the robustness of the autonomous robotic system and increases the task success rate.

In this paper, we present a language model-based framework enabling the autonomous execution of loco-manipulation tasks. This framework leverages the semantic understanding capabilities of LLM in reaction to human instructions, which is based on the knowledge of the task scenario and the behavioral skills (“action" and “perception") possessed by the robot. As shown in Figure 1, when a human provides the instruction “find the bottle and pick it up", the LLM will generate a $\textit{task graphs}$ consisting of different behavioral nodes based on the instruction and the behavior lib as prompts, which will guide the CENTAURO robot \cite{8630605} to perform the actions such as object detection, object grasping, and lifting, etc. While the VLM-based perceptual behavior node is used as a failure detector, it's triggered when the state of the current task is required to be detected and makes an inferential judgment based on the images returned from the robot onboard camera.

Our summary of the main contributions of this work includes:
\begin{itemize}
\item We exploit the appropriateness of LLM in loco-manipulation tasks and propose a novel framework for autonomous behavior planning, enabling rapid deployment without additional training, which can be applied to quadrupeds, humanoids, and mobile manipulators. 
\item We propose a new paradigm for robot behavior library, which enables to encoding of human instruction into optimal action sequences by combining behaviors in a modular way and linking high-level tasks to interpretable low-level control.
\item We incorporate multimodal failure detection as higher-order feedback to facilitate the task graph in correcting the misalignment between the intended goal and the actual robot's actions during the execution of the task.
\end{itemize}

 The proposed framework was experimentally validated to assess its feasibility in handling long-horizon tasks in simulation and real-world environments.
 

\section{Related works}
Endowing robots with autonomy in locomotion and manipulation tasks still represents a high-level challenge for robotics. In recent years, prior research has focused on motion planning and trajectory optimization \cite{humanoidmpc}, covering systems with various morphologies and levels of autonomy. The data-driven approach \cite{kappler2015data} enables the use of experience to make decisions online and generate appropriate multimodal reference trajectories for dexterous manipulation. Meanwhile, combining data and learning, the model-free reinforcement learning approach \cite{9684679}\cite{sleiman2023versatile} has demonstrated impressive performance in several specific tasks and unstructured environments. For long-term tasks, \cite{10243138}\cite{9359455} introduced a novel motion planning framework and task evaluation approach that allowed robots to maintain dexterity while navigating complex environments. Boston Dynamics \cite{boston} utilized an offline trajectory optimization and model predictive control strategy to codify the Atlas robot's movements into a time series, achieving consistency between simulation planning and motion execution. However, when faced with different task scenarios and long-horizon planning involving multiple subtasks, the ability to understand instructions and reason about tasks is often overlooked, making it challenging to achieve autonomy and adaptability in task-driven mobile manipulation.

With the emergence of large language models (LLMs), several transformer-based architectural planners \cite{brohan2022rt}\cite{brohan2023rt}\cite{padalkar2023open} have played a pivotal role in predicting and generating robot actions and attempting to derive robot policy code \cite{liang2023code} guided by natural language instructions. However, such end-to-end strategies often require substantial amounts of training data and expert demonstrations, complicating the model's training process, especially in unknown scenarios. In contrast, some approaches leverage LLMs' semantic comprehension capabilities as a higher-level task planner \cite{ahn2022can}\cite{huang2022language}\cite{ding2023task}, transforming instructions into executable lower-level actions in a zero- or few-shot manner \cite{huang2023voxposer}. Nonetheless, these methods tend to assume the success of the task performed and overlook the potential discrepancy between planned expectations and real-world execution. They often fail to enable whole-body control in multi-joint, high degree-of-freedom (DoF) mobile manipulation tasks, such as those involving humanoid robots. Moreover, studies such as \cite{huang2023voxposer}\cite{driess2023palm}\cite{yang2023vid2seq} attempt to interpret textual and visual inputs simultaneously, using them to address downstream robotic tasks. While visual question answering (VQA) \cite{antol2015vqa} can be effectively achieved through understanding image descriptions and inferring the robot's state and current context to guide subsequent actions, relying solely on high-level visual feedback has been proved to be insufficient. It compromises the rapid response to dynamic environments and is less effective than other methods exploring proprioceptive perception, such as interaction force sensing when executing low-level tasks. 

Our study, on the other hand, applies the LLM to robots by utilizing its ability to understand instructions and plan with existing behavior libraries by generating task graphs that sequentially instruct the high DoF humanoid robot's whole-body actions. It detects and recovers from possible failures during tasks by using the VLM as a perceptual behavioral node and combining it with other perceptrons to enable multimodal feedback. This approach allows the robot to serialize discrete action nodes and select different perceptual combinations according to the task scenario. Ultimately, a robust autonomous mobile manipulation skill is realized and demonstrated.

\section{Autonomous Robot Behavior Planning}

\subsection{Problem Statement} 
The problem of performing autonomously loco-manipulation task based on higher-level instructions can be described as follows.
We assume that the human's semantic description of the task, denoted as $i$ articulates a specific task to be executed by the robot. 
Additionally, we consider that a $\textit{behavior lib}$ $\Pi$ is provided to the robot, consisting of a set of action and perception behaviors $\pi \in \Pi$ that can be directly executed by the robot, such as the grasp action would enable the robot to control the gripper for grasping in Cartesian space, while the object detection behavior would utilize the camera to detect the pose of the target object. Each behavior is associated with a specific semantic description $l_{\pi}$ (e.g., “open the griper and move to the target pose then close griper"). By invoking the large language model $L$, the task graph $T$ corresponding to instruction $i$ is generated:
\begin{equation}
    T_{i} = L(i, l_{\pi_{1}}, l_{\pi_{2}},...,l_{\pi_{n}}) 
\end{equation}
The task graph $T_{i}$, encapsulates a sequential arrangement of behaviors necessary for the robot to execute in various states $s$ to accomplish the designated task requested by the human. This graph is maintained in an XML file format and is operationalized through the Behavior Tree (BT) $B$. The robot state is the feedback by the execution results of different behavioral nodes, including (running, success, fail). Depending on the current state of the robot, the BT guides the robot to execute different behaviors. When a low-level execution deviation from the task plan is detected, the BT fixes this failure by resuming the behavior in an attempt to correct the error. 

The above process is described in Algorithm 1. In this way, the robot is enabled to encode human instructions into various sequences of behaviors and execute them according to the task demands, as well as to recover possible misalignment between the instructions and the robot's execution.

\begin{algorithm}
\caption{Language model based robot behavior planner}\label{alg}
\hspace*{\algorithmicindent} \textbf{Given}: Language model $L$, a human instruction $i$, and a behavior lib $\Pi$ with the language description $l_{\pi}$.
\begin{algorithmic}[1]
\STATE$T_{i}$ = $L(i, l_{\pi})$
\STATE$B \gets BehaviorTree(T_{i})$
\STATE{Initialize the state $s_{\pi}$, number of steps $n$}
\WHILE{$B(s_{\pi_{n}}) \ne "done"$}
\FOR{$\pi_{n} \in \Pi_{B}$}
\STATE{executeBehavior$\pi_{n}$}
\STATE$s_{\pi_{n}} \gets update Status$
\ENDFOR
\IF{allBehavior Completed}
\RETURN{Done}
\ENDIF
\ENDWHILE
\end{algorithmic}
\end{algorithm}

\subsection{Overview of the framework}

\begin{figure*}
 \centerline{\includegraphics[width=16cm]{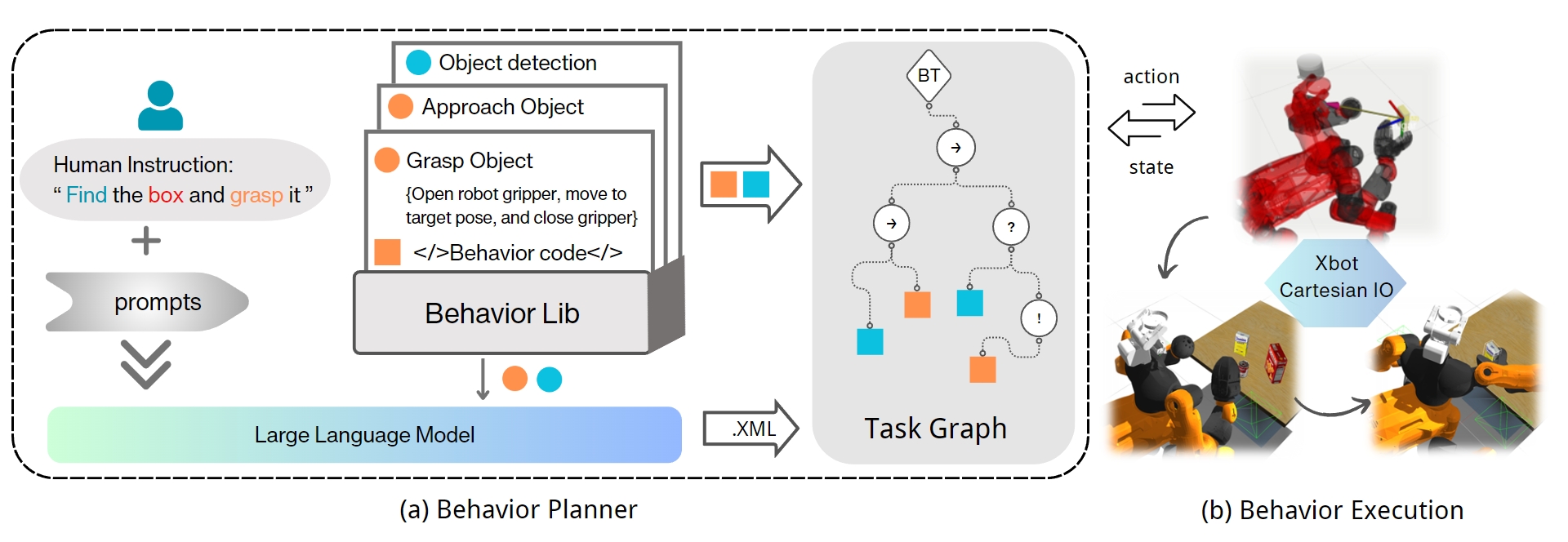}}
\caption{Overview of the Framework. (a) Behavior Planner takes the human instruction as input, given the behavior lib and prompts, LLM generates a hierarchical structure behavior tree, which forms the task graph along with the behavior code. (b) The CENTAURO robot executes the lower action command and feeds back its current state. The entire process does not require any additional training. }
\label{framework}   
\end{figure*}

We propose a robot behavior planning system for performing autonomously loco-manipulation tasks that makes use of language models. Figure 2 shows an overview of the system. We first create a library of behaviors for the CENTAURO robot, divided into action behaviors and perceptual behaviors, each containing a corresponding behavior tag and a behavior code. The behavior tag records the name and type of the behavior, as well as a detailed semantic description of the behavior. In addition, prompts are defined in advance. Prompts include conditional information for input to the large language model, a description of the robot's features, the skills that the robot possesses ($\textit{behavior lib}$), and a description of the expected outputs. When a robot is called upon to perform a task, the human first gives task instructions as input, which are fed to the large language model together with the prompts, along with the behavior tags from the $\textit{behavior lib}$. Upon obtaining this information, the LLM generates a sequence comprising the robot's behaviors based on the given task and stores it in an XML-formatted file, which is used to generate a Behavior Tree (BT) \cite{colledanchise2018behavior} that controls the robot's task execution. The behavior code in $\textit{behavior lib}$ and the BT generated by the LLM together form a $\textit{task graph}$, which instructs the robot to perform different behaviors according to the current robot state and conditions, and ultimately completes the task. The task graph also plans a correction policy for failures during the task. If the perceptual behavior detects an inconsistency in the execution of the task, the task graph will execute specific behaviors to try to fix the failure.

While in the behavior execution phase, the CENTAURO depends on Xbot\cite{muratore2020xbot} and Cartesian I/O\cite{laurenzi2019cartesi} to execute the action commands issued by the task graph, the current state of the robot, as well as the sensory information, are fed back to the task graph to guide the next action. During this process, the host PC is responsible for behavioral planning, including receiving human instructions and behavior tags, and generating the task graph through the large language model. Meanwhile, the robot pilot is tasked with receiving and executing the low-level action commands.

\subsection{Language Model for Behavior Planning and Modification}
\subsubsection{Behavior Lib}
Controlling a robot to accomplish complex actions in a long-horizon task is difficult and challenging. To link semantic behaviors with the actual execution of actions by the robot, we designed a behavior library for the CENTAURO robot such that each skill can directly control the robot to complete basic actions. This approach improves the interpretability of each step of the task process and reduces the deviation between high-level tasks and low-level execution by partitioning the complex task into a sequence of actions consisting of several behaviors skills. 

We classify the $\textit{behavior lib}$ into action behaviors and perceptual behaviors. Action behaviors control the CENTAURO robot to complete whole-body motions such as moving and manipulating. Inspired by object manipulation in daily office environments, we have designed several individual robot actions to compose the action lib using Cartesian I/O\cite{laurenzi2019cartesi}, including \texttt{Approach}, \texttt{Grasp}, \texttt{Lift}, etc., which are used to implement subtasks such as navigating to various locations, grasping a target object, and lifting a target object, etc. The action lib is a library of robot actions that can be added and combined depending on the demands of a loco-manipulation task and the interaction environment. Meanwhile, perceptual behaviors rely on the robot's internal sensors ( torque sensing, RGBD camera) to detect the position of the object, evaluate the robot's state, and reason whether the current task is complete or if there are failures. Various algorithms \cite{tremblay2018deep}\cite{gpt4v} are integrated into different perceptual behaviors, and similar to action behaviors, perceptual behaviors can be designed independently and added to the perceptual lib. Multiple sensors can be invoked in a single perceptual behavior and fused with data from different modalities according to the requirements.

\begin{table}[h]
\caption{Behavior Lib Definition}
\centering
\begin{tabular}{lll} 
\toprule
\textbf{Behavior} & \textbf{Type} & \textbf{Tag}                                                                                           \\ 
\hline
\texttt{Homing}   & $action$        & \begin{tabular}[c]{@{}l@{}}‘bringing all of the joints of robot \\to homing configuration’\end{tabular}  \\ 
\hline
\texttt{Approach}          & $action$        & \begin{tabular}[c]{@{}l@{}}‘moving robot torso closer \\to target by certain distance’\end{tabular}    \\ 
\hline
\texttt{Grasp}             & $action$        & \begin{tabular}[c]{@{}l@{}}‘moving gripper to \\a given pose and close it’\end{tabular}                \\ 
\hline
\texttt{Lift}              & $action$        & \begin{tabular}[c]{@{}l@{}}‘raising gripper to \\the chest and adjusting pose’\end{tabular}            \\ 
\hline
\texttt{Place}             & $action$        & \begin{tabular}[c]{@{}l@{}}‘moving gripper to \\the given position and open it’\end{tabular}           \\ 
\hline
\texttt{Distance}          & $perception$    & \begin{tabular}[c]{@{}l@{}}‘measuring distance\\between object and robot’\end{tabular}                 \\ 
\hline
\texttt{Grip force}        & $perception$    & \begin{tabular}[c]{@{}l@{}}‘obtaining the actual \\torque of gripper’\end{tabular}                     \\ 
\hline
\texttt{Object detection}  & $perception$    & \begin{tabular}[c]{@{}l@{}}‘detecting and estimating \\6-DoF of objects’\end{tabular}                  \\ 
\hline
\texttt{Visual Q\&A}               & $perception$    & \begin{tabular}[c]{@{}l@{}}‘reasoning task state \\using visual language model’\end{tabular}                    \\
\bottomrule
\end{tabular}
\end{table}

\subsubsection{LLM Generated Task Planner}
While large language models can utilize their extensive knowledge of semantic data as well as their text comprehension reasoning capabilities to provide answers to human instructions, the answers can be diverse. To obtain the desired output, it is necessary to impose constraints on the instructions given as input. One approach is to use prompt words, a linguistic construct designed to qualify a language model to give a specific output. In our framework, prompts are used as input to the large language model along with human instructions and behavior tags, which consist of several components. First, there is information about the current state of the CENTAURO robot and its hardware configuration. Then, the concept of a behavior library, its components, and sample applications are introduced. Lastly, there is the expected output in a format that includes the concept of a behavior tree, the definition of the nodes, and examples of applications.

\begin{figure}
 \centerline{\includegraphics[width=8cm]{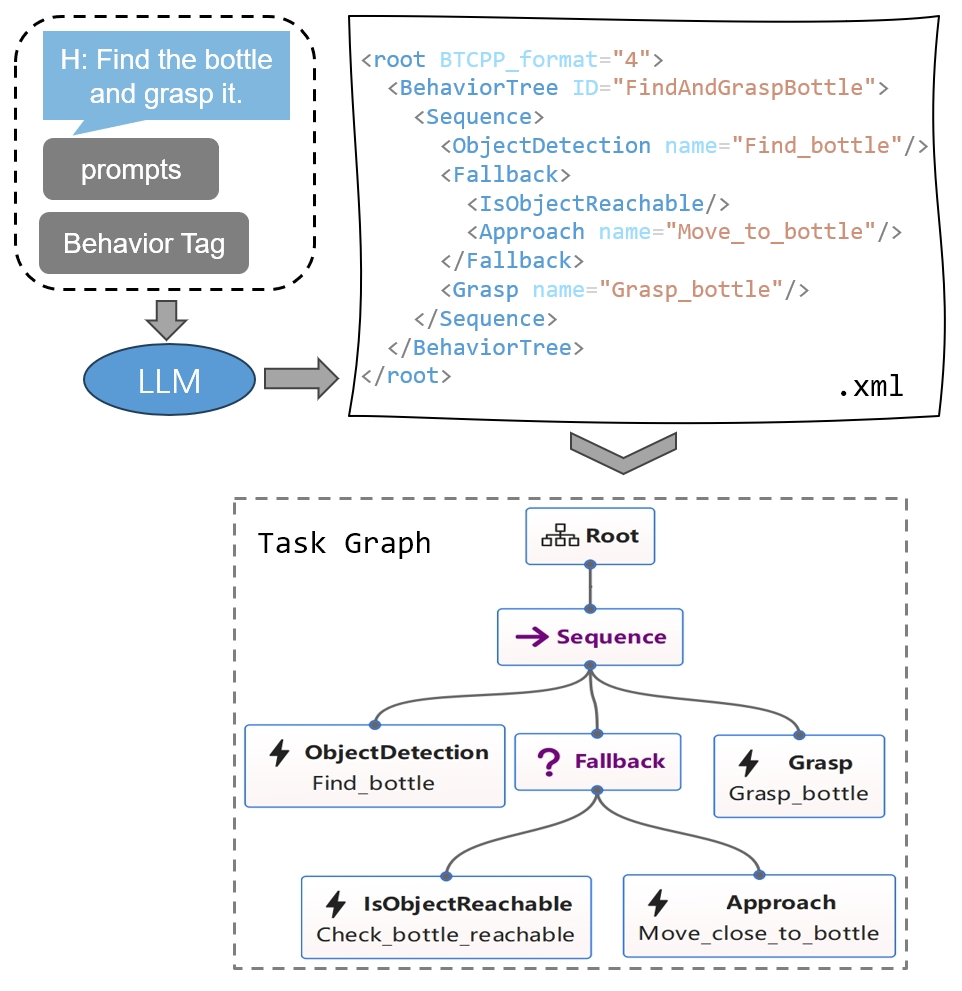}}
\caption{Behavior Planner Grounding LLM}
\label{planner}   
\end{figure}

To convert high-level instructions into a sequence of implementable low-level skills, we leverage Behavior Trees as both an intermediate bridge and an output of the LLM. The use of Behavior trees provides a hierarchical, tree-structured framework for controlling the robot's actions and decision-making processes \cite{iovino2022survey}. 
This framework consists of nodes with different functions, including those controlling the execution process and conditional judgments, as well as nodes that actually execute the robot's actions. Having previously defined the $\textit{behavior lib}$, the LLM generates the behavior tree framework based on the behavioral skills and task instructions. This framework is stored in an XML file. The $\textit{task graph}$ is responsible for loading the behavior tree and invoking behaviors from the $\textit{behavior lib}$ according to the node guidance. This setup realizes decision reasoning through the LLM. The $\textit{task graph}$ is used for behavior planning, and the robot ultimately executes the tasks.

We access the \texttt{gpt-4} model as the LLM through the OpenAI API, which directly outputs the XML file used for generating the behavior tree, as shown in Figure 3.

\subsubsection{Failure Detection and Recovery}
In order to determine whether a task is successfully completed or deviates during execution, we try to incorporate a failure detection and recovery mechanism into the task graph. In our work, to take advantage of the visual language model's capability of understanding and reasoning about images, we utilize visual questions and answers (VQA) as perceptual behaviors to determine the current state of the robot performing the task, such as in the task of ‘picking the box’ by giving the robot's camera image and asking “Is the box being held?”, the VLM will respond to the query by answering {“\texttt{Yes} ” or “\texttt{No}”}. Proprioceptive sensing like torque and distance has also been developed as behaviors to detect the potential failures in specific tasks, like during the tasks requiring grasping, detecting the torque on the gripper can be a reference of whether the object is being held. The perceptual behaviors we define in the behavior lib give multiple alternatives and combinations for the failure detection nodes, allowing the LLM to design the behavior tree based on the reasoning of different tasks. Some simple tasks such as "find and approach to object" only require the initiation of \texttt{Object detection} behavior to determine if the object is available, while tasks that require multiple robotic actions often demand a combination of different perceptual behaviors for failure detection.

During the execution of the behavior tree, the node will return three signals: $success$, $failure$, and $running$, and the behavior tree will guide the execution of the behavior according to the returned signals. After the action node is executed, the condition node can be added to decide whether the current task is successful or returns the current state of the target object. For instance, in the process of grasping and lifting an object, after the completion of grasping, a condition node \texttt{IsObjectHeld} can be added to decide whether the object has been successfully grasped or not. In this scenario, the node will activate the \texttt{Grip force} and \texttt{Visual Q\&A} perception behavior, which will obtain the torque of the gripper and ask the VLM "Is the (Target Object) held by the gripper". Only if there is a torque on the gripper and the VLM answers "Yes", then the node will return a success signal. The behavior tree will continue to execute the subsequent nodes. If it returns a failure, the recovery node is activated and the robot will try to grasp the object again. 
\begin{figure}
 \centerline{\includegraphics[width=8cm]{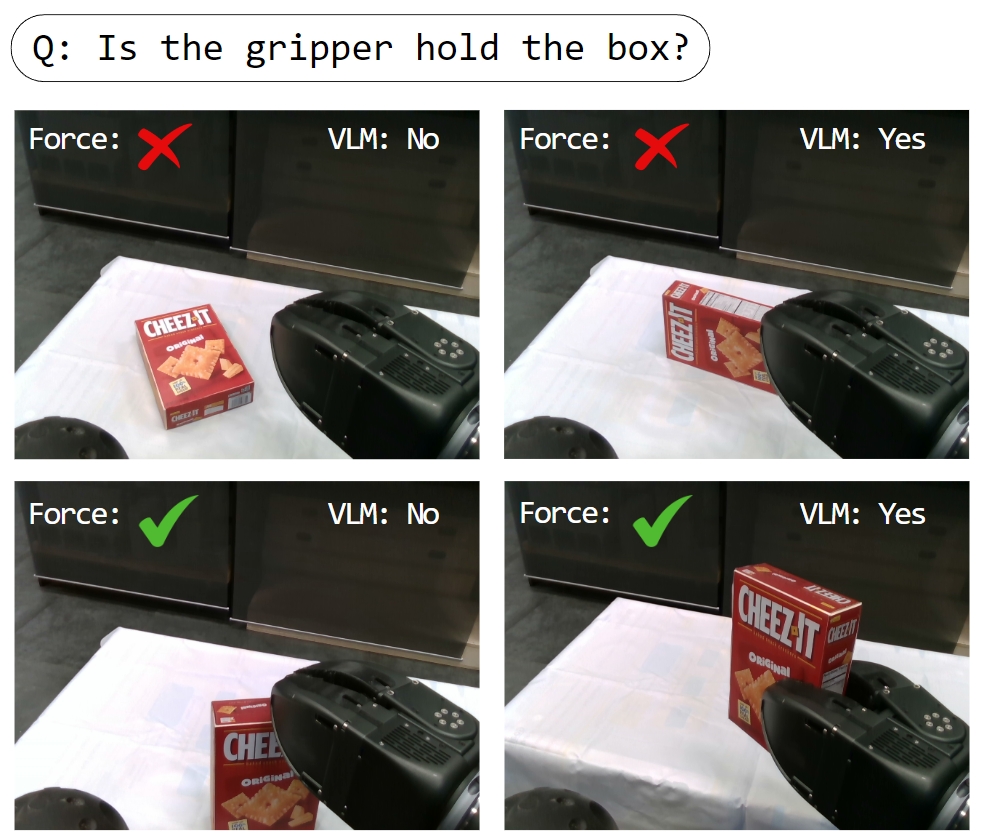}}
\caption{Failure detection using a combination of perception behaviors. By asking VLM, the \texttt{visual Q\&A} behavior can reason the state of the task, while using the torque sensor, the \texttt{Grip force} behavior will return the torque on the gripper.}
\label{failure}   
\end{figure}

\section{Experiment and Evaluation}
We experimentally verify the capability of LLM as a behavior planner by implementing it and assessing its performance on CENTAURO robot executing long-horizon tasks under semantic commands. Few studies have employed LLM to plan the behavior of humanoid type of robots like CENTAURO and conducted real-world experiments. It is challenging to use different robots as a control group due to the variations in their functionalities and configurations. Therefore, we compare this method with our previous study in terms of functional aspects as shown in Table 2, and conduct preliminary experiments on applying LLM on the CENTAURO robot.

\subsection{Experiment Setup}

We conducted the experiment using objects from the YCB dataset \cite{calli2015benchmarking} that are commonly found in an office kitchen. The test environment was an open area inside the lab, with objects randomly placed on a desk as shown in Figure 5. The CENTAURO robot, a hybrid wheels and legs quadruped robot with a humanoid upper body, features 37 degrees of freedom and a two-fingered claw gripper, enabling it to perform a wide range of loco-manipulation tasks. Equipped with an RGBD camera on its head and torque sensors in the joints throughout its body, the robot possesses extensive perceptual capabilities to measure joint efforts and interaction forces.

The action behaviors in the behavior lib were designed based on Cartesian I/O, requiring no extra training. The Xbot functions as middleware, providing real-time communication between the robot's various underlying actuators and the task commands through an API interface. For message transmission between the LLM, behavior lib, Behavior Tree, and the robot, we utilized the Robot Operating System (ROS) and conducted simulation experiments in the Gazebo simulator. Experiments for both simulation and the real world are demonstrated in the accompanying video.

\begin{figure}
 \centerline{\includegraphics[width=6cm]{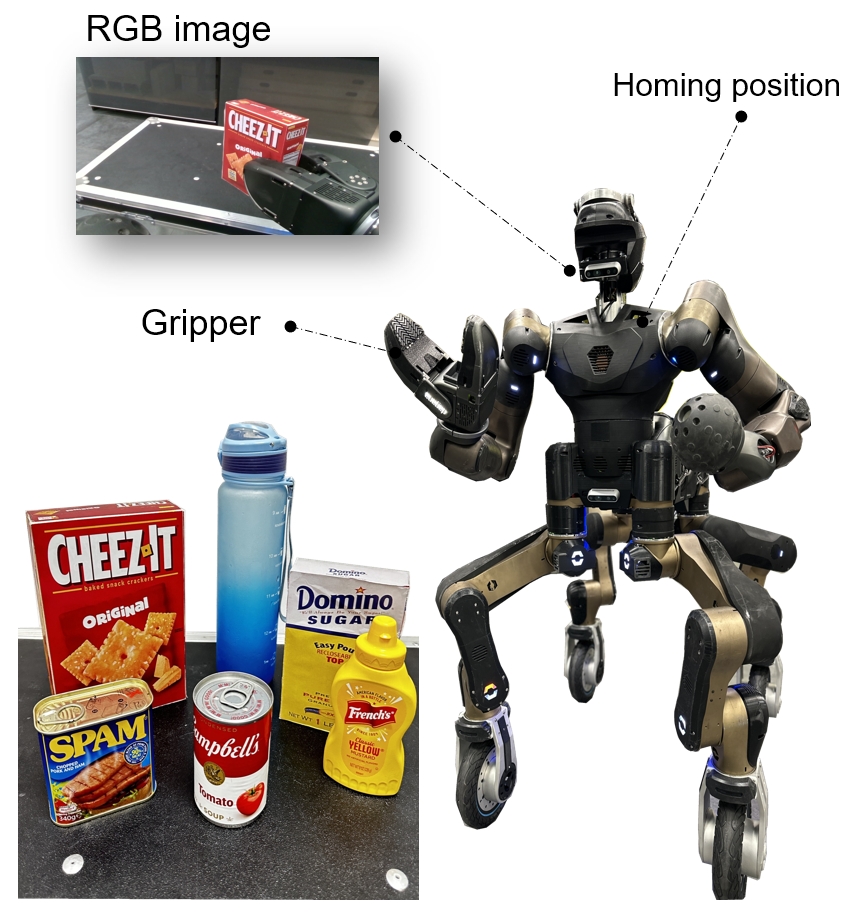}}
\caption{Experiment setup}
\label{setup}   
\end{figure}

\subsection{Autonomous Humanoid Loco-manipulation Task}

\begin{table}
\centering
\caption{Comparison of different methods}
\begin{tblr}{
  row{1} = {c},
  row{2} = {c},
  cell{1}{1} = {r=2}{},
  cell{1}{2} = {c=2}{},
  cell{3}{2} = {c},
  cell{3}{3} = {c},
  cell{4}{2} = {c},
  cell{4}{3} = {c},
  cell{5}{2} = {c},
  cell{5}{3} = {c},
  cell{6}{2} = {c},
  cell{6}{3} = {c},
  cell{7}{2} = {c},
  cell{7}{3} = {c},
  hline{1,8} = {-}{0.08em},
  hline{2} = {2-3}{},
  hline{3-7} = {-}{},
}
\textbf{Abilities}   & \textbf{Method}            &                             \\
                     & {Whole-body MPC\\ \cite{humanoidmpc}}    & {LLM Behavior Planner\\(ours)} \\
Autonomy             & low                        & high                        \\
Whole-body motion    & \ding{52}                  & \ding{52}   \\
Long-horizon task    & \ding{55}                  & \ding{52}   \\
Failure detection    & \ding{55}                  & \ding{52}   \\
Real-time replanning & \ding{52}                  & \ding{55}   
\end{tblr}
\end{table}

\subsubsection{Behavior Planning with LLM}

We first tested the LLM's behavior planning capabilities for robot tasks of varying complexity. The experiment was conducted for eight different tasks, including tasks with failure detection and recovery (FR), and the behaviors were planned using the method shown in Figure. 3. We provided standard instructions only for the given objects in Figure. 5, and the behaviors created in the behavior lib. These instructions are simple descriptions of the task content, e.g., “Find the soup can and pick it up." If failure detection and recovery are required during the task, this will need to be stated in the instructions such as “Pick the cracker, place it aside. Detect and recover the failure during the task.". We then used the Behavior Tree to load the XML files generated by the LLM and verified their feasibility. Finally, the appropriateness of the behavioral planning and the successful completion of the task were manually verified. Experiments that were executed and followed the requirements of the task instructions for planning were judged as successful. Each task was planned a total of 50 times, all using the same behavior lib and prompt. The time for each task graph generation was recorded, as well as the executable and success rate of the behavioral planning, as shown in Table 3.

\begin{table}
\centering
\caption{Behavior planning results for different tasks including tasks with failure detection and recovery (FR).}
\begin{tblr}{
  column{4} = {c},
  cell{1}{1} = {c},
  cell{2}{2} = {c},
  cell{2}{3} = {c},
  cell{3}{2} = {c},
  cell{3}{3} = {c},
  cell{4}{2} = {c},
  cell{4}{3} = {c},
  cell{5}{2} = {c},
  cell{5}{3} = {c},
  cell{6}{2} = {c},
  cell{6}{3} = {c},
  cell{7}{2} = {c},
  cell{7}{3} = {c},
  cell{8}{2} = {c},
  cell{8}{3} = {c},
  cell{9}{2} = {c},
  cell{9}{3} = {c},
  hlines,
  hline{1,10} = {-}{0.08em},
}
\textbf{Task}              & \textbf{Executable} & \textbf{Success} & \textbf{Time(s)} \\
Find object                & 100\%               & 94\%             & 14.93            \\
Approach object            & 98\%                & 90\%             & 16.15            \\
Grasp object               & 96\%                & 92\%             & 16.27            \\
Pick object                & 96\%                & 84\%             & 17.11            \\
Pick object (FR)           & 90\%                & 82\%             & 17.91            \\
Pick and place object      & 92\%                & 84\%             & 18.23            \\
Pick and place object (FR) & 84\%                & 80\%             & 19.07            \\
Find and pick object (FR)  & 86\%                & 82\%             & 17.86            
\end{tblr}
\end{table}

\begin{figure}
     \centerline{\includegraphics[width=8cm]{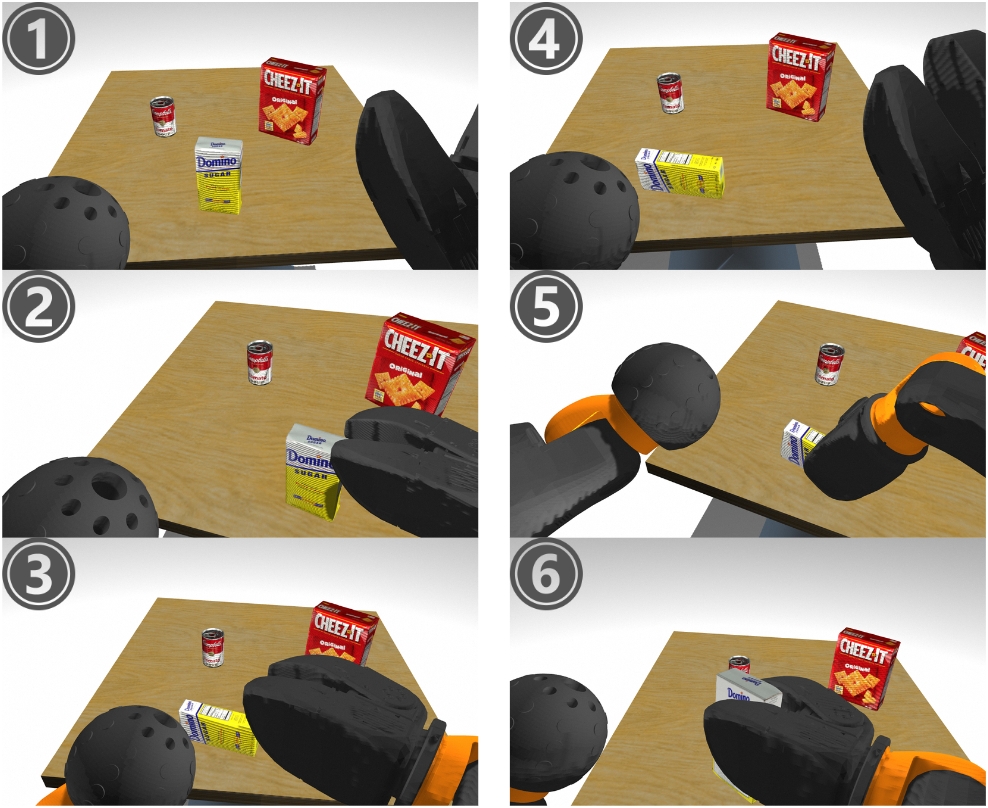}}
    \caption{Task execution with failure detection and recovery in simulation. Images 1, 2, and 3 show the robot's first attempt to pick up an object. After the perception behaviors detected that the gripper did not successfully grasp the object in image 3, then the robot tried again and successfully picked the object as shown in image 4, 5, 6.}
    \label{simulation}
\end{figure}

\begin{figure*}
     \centerline{\includegraphics[width=18cm]{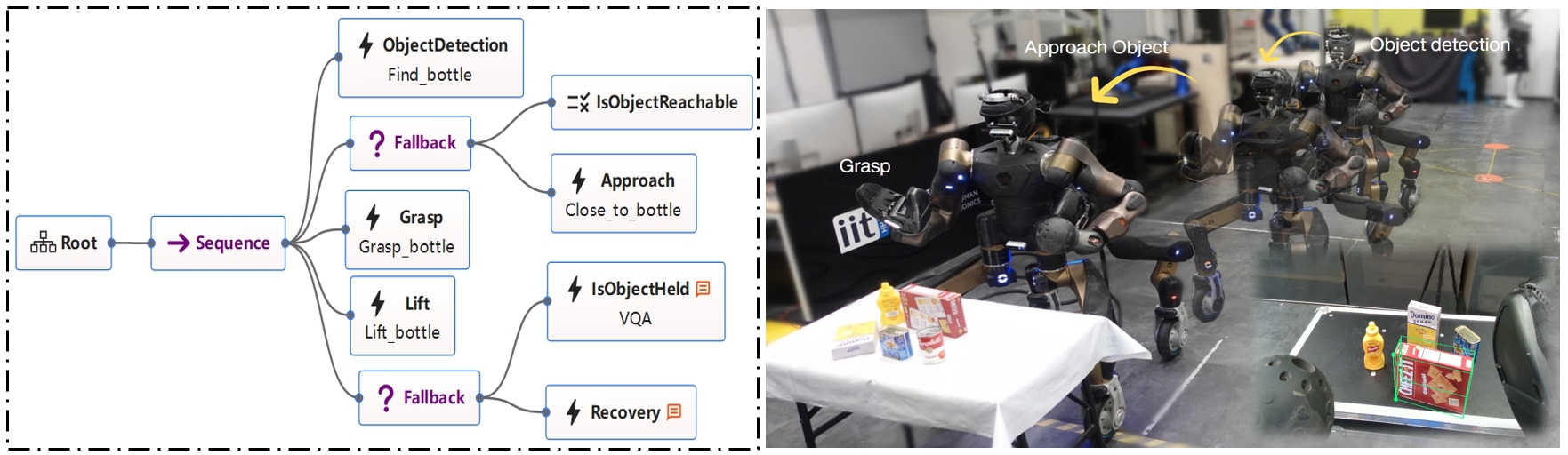}}
    \caption{Long-horizon task planning with CENTAURO robot. The left shows the task graph generated using LLM, the right shows the robot executing human instruction according to the behavioral plan.}
    \label{realworld}
\end{figure*}

\subsubsection{Long-horizon Task Execution}

After verifying that the behavioral plans generated by the LLM can be converted into an executable Behavior Tree, we conducted experiments using the CENTAURO robot in both simulation and real-world environments. We selected the last six tasks from Table 3 to test the actual performance of the robot executing LLM-planned behaviors. Starting with object grasping and picking tasks, the initial positions of the robot were set at different distances from the target objects, which were randomly placed on the table. In the last task, the CENTAURO was placed relatively far to test the performance of the planner in tasks where “approach" behavior was necessary. For the same task, different descriptions of instruction and different target objects were used to verify the LLM's ability to reason about the simple task and plan the robot's behavior. For tasks that require in-process failure detection and recovery, the LLM incorporates perceptual behaviors in the behavioral planning phase and attempts to recover if it detects that the planned action fails to complete the task. This is shown in Figure. 6, where the recovery behaviors are selected based on the task requirements and the current state of the robot. Finally, the experiments focus on verifying the LLM's behavioral planning for long-horizon tasks and the robot's ability to perform autonomous loco-manipulation, with multiple perceptual and action behaviors being selected and combined to achieve the task goal, as shown in Figure. 7. We conducted 25 experiments for each task separately and recorded the success rate and execution time of the robot to complete the task in both simulation and the real environment, as shown in Table 4.

\begin{table}
\centering
\caption{Experiment results of simulation and real-world environment}
\begin{tblr}{
  row{1} = {c},
  row{2} = {c},
  cell{1}{1} = {r=2}{},
  cell{1}{2} = {c=2}{},
  cell{1}{4} = {c=2}{},
  cell{3}{2} = {c},
  cell{3}{3} = {c},
  cell{3}{4} = {c},
  cell{3}{5} = {c},
  cell{4}{2} = {c},
  cell{4}{3} = {c},
  cell{4}{4} = {c},
  cell{4}{5} = {c},
  cell{5}{2} = {c},
  cell{5}{3} = {c},
  cell{5}{4} = {c},
  cell{5}{5} = {c},
  cell{6}{2} = {c},
  cell{6}{3} = {c},
  cell{6}{4} = {c},
  cell{6}{5} = {c},
  cell{7}{2} = {c},
  cell{7}{3} = {c},
  cell{7}{4} = {c},
  cell{7}{5} = {c},
  cell{8}{2} = {c},
  cell{8}{3} = {c},
  cell{8}{4} = {c},
  cell{8}{5} = {c},
  hline{1,9} = {-}{0.08em},
  hline{2} = {2-5}{},
  hline{3-8} = {-}{},
}
\textbf{Task}              & \textbf{Simulation} &         & \textbf{Real World } &         \\
                           & Success             & Time(s) & Success              & Time(s) \\
Grasp object               & 92\%                & 85.7    & 96\%                 & 98.4    \\
Pick object                & 84\%                & 104.9   & 80\%                 & 121.3   \\
Pick object (FR)           & 88\%                & 116.2   & 88\%                 & 167.1   \\
Pick and place object      & 76\%                & 132.7   & 72\%                 & 160.6   \\
Pick and place object (FR) & 84\%                & 189.2   & 80\%                 & 203.2   \\
Find and pick object (FR)  & 80\%                & 174.5   & 76\%                 & 197.8   
\end{tblr}
\end{table}

\subsection{Results analysis}
In the experiments, we evaluated the behavioral planning capabilities of the LLM for tasks with varying complexity levels, applying it to the CENTAURO robot. With a defined behavior lib and appropriate prompts, the LLM can generate corresponding behavior plans based on different task instructions, achieving a high planning success rate and task execution rate. These rates vary with the task's complexity and the number of behaviors needed to complete it. By comparing the original tasks with the tasks including FR, incorporating failure detection and recovery into the task process increases the difficulty of behavior planning, affecting the success rate of the generated task graphs. By adding FR, it increased the time for task execution with the robot as shown in Table 3, but not significantly increase the time for LLM planning. Finally, the behavioral planning time depends on the feedback speed of the language model used and the hardware device response time. The task complexity primarily affects the time taken to load the Behavior Tree, leading to minor differences in planning time across different tasks.

The results from robot task execution in both simulation and real environments demonstrate that LLM can effectively plan humanoid robot loco-manipulation tasks to a considerable degree. By integrating perception and action behavior in the behavior lib through LLM, the CENTAURO robot reaches a satisfactory level of success rate ($\ge 72\%$) in task execution. In long-horizon tasks, the incorporation of failure detection and recovery significantly boosts the robot's execution success rate in both simulation and real-world settings, and the success rate can be increased by up to $8\%$ in specific tasks. Additionally, increasing task complexity and the addition of more robot behavioral nodes result in extended task implementation times as shown in Table 4.

\section{Conclusion}

In this work, we introduce an autonomous online behavioral planning framework utilizing a large language model (LLM) for performing robot loco-manipulation tasks, requiring only human language instructions. Within this framework, we propose the concept of a behavior library and design action and perception behaviors, which are both interpretable and pragmatically efficient, with corresponding behavioral tags provided for semantic interpretations. The LLM organizes these behaviors into a task graph with a hierarchical structure, derived from the understanding of given instructions. The robot then follows the nodes in this task graph to sequentially complete the task. Additionally, it detects and attempts to correct possible failures by integrating the visual language model with intrinsic perceptions throughout the task process, thus successfully planning and executing long-horizon tasks. Experiments with the CENTAURO robot validate the achieved performance and practicality of this framework in robotic task planning.

Future work will focus on enriching the robot's behavior lib, as well as improving the prompts system, so that the LLM can better plan and optimize behavioral sequences automatically based on the robot's intrinsic mobility, manipulation, and perceptual strengths, thus enabling to perform more complex mobile manipulation tasks. Another direction is to improve the dynamic planning and multiconditional reasoning capability of the framework. This includes behavioral replanning in response to external perturbations or the introduction of artificial subtasks during a task.

\bibliographystyle{IEEEtran}
\bibliography{references}

\addtolength{\textheight}{-12cm}   

\end{document}